# An Immune System Inspired Approach to Automated Program Verification

Soumya Banerjee

*Abstract* — An immune system inspired Artificial Immune System (AIS) algorithm is presented, and is used for the purposes of automated program verification. Relevant immunological concepts are discussed and the field of AIS is briefly reviewed. It is proposed to use this AIS algorithm for a specific automated program verification task: that of predicting shape of program invariants. It is shown that the algorithm correctly predicts program invariant shape for a variety of benchmarked programs.

## I. INTRODUCTION

OVER the last few years, there has been an ever-increasing interest in the area of artificial immune systems (AIS) and their applications [1]–[6]. AIS uses ideas gleaned from immunology in order to develop adaptive systems capable of performing a wide range of tasks in various areas of research.

In this paper, I review the clonal selection concept, together with the affinity maturation process, and state how these biological principles can lead to the development of useful computational tools [7]. The algorithm proposed to be used focuses on a systemic view of the immune system and does not take into account cell–cell interactions. I do not model any biological phenomenon, but propose how some basic immune principles can help us to not only better understand the immune system itself, but also to solve a problem in program verification: that of finding a program invariant.

An *invariant* of a program is a mathematical formula that captures the semantics of the program [8] and is used in automatic program verification. The *shape* of an invariant is its approximate polynomial representation. Once the shape of the invariant is predicted, deterministic techniques can be used to generate the exact form of the invariant [9]. Hence, the prediction of invariant shape is of paramount importance for program verification.

An AIS algorithm is proposed to carry out the machine-learning task of predicting invariant shape from an instance of a program. No distinction is made between a B cell and its receptor, known as an antibody (*Ab*), so that every element of our artificial immune system will be generically called an *Ab*.

## II. CLONAL SELECTION THEORY

Any molecule that can be recognized by the adaptive immune system is known as an antigen (*Ag*). When an animal is exposed to an *Ag*, some subpopulation of its bone-marrow-derived cells (B cells) responds by producing antibodies (*Ab*'s). *Ab*'s are molecules attached primarily to the surface of B cells whose aim is to recognize and bind to *Ag*'s. Each B cell secretes a single type of *Ab*, which is relatively specific for the *Ag*. By binding to these *Ab*'s and with a second signal from accessory cells, such as the T-helper cell, the *Ag* stimulates the B cell to proliferate (divide) and mature into terminal (nondividing) *Ab* secreting cells, called plasma cells. The process of cell division (mitosis) generates a clone, i.e., a cell or set of cells that are the progenies of a single cell.

B cells, in addition to proliferating and differentiating into plasma cells, can differentiate into long-lived B memory cells. Memory cells circulate through the blood, lymph, and tissues and, when exposed to a second antigenic stimulus, commence to differentiate into plasma cells capable of producing high-affinity Ab's, preselected for the specific Ag that had stimulated the primary response. The main features of the clonal selection theory [10], [11] that will be reviewed in this paper are:

   1) proliferation and differentiation on stimulation of cells with Ag's;
   2) generation of new random genetic changes, expressed subsequently as diverse Ab patterns, by a form of accelerated somatic mutation (a process called affinity maturation);

### A.    Reinforcement Learning and Immune Memory

Learning in the immune system involves raising the relative population size and affinity of those lymphocytes that have proven themselves to be valuable by having recognized a given Ag. In the use of the clonal selection theory for the solution of practical problems, I do not intend to maintain a large number of candidate solutions, but to S.B. Author is with the Department of Computer Science, University of New Mexico, Albuquerque, NM - 87106, USA. (Phone - 505-277-3122, Fax - 505-277-6927, E-mail: soumya@cs.unm.edu)

keep a small set of best individuals. A clone will be created temporarily and those progeny with low affinity will be discarded. The purpose is to solve the problem using a minimal amount of resources. Hence, I seek high-quality and parsimonious solutions.

In the normal course of the immune system evolution, an organism would be expected to encounter a given *Ag* repeatedly during its lifetime. The initial exposure to an *Ag* that stimulates an adaptive immune response is handled by a small number of low-affinity B cells, each producing an *Ab* type of different affinity. The effectiveness of the immune response to secondary encounters is enhanced considerably by the presence of memory cells associated with the first infection, capable of producing high-affinity *Ab*'s just after subsequent encounters. Rather than "starting from scratch" every time, such a strategy ensures that both the speed and accuracy of the immune response becomes successively higher after each infection.

This is an intrinsic scheme of a *reinforcement learning strategy* [12], where the interaction with the environment gives rise to the continuous improvement of the system capability to perform a given task. To illustrate the adaptive immune learning mechanism, consider that an antigen $Ag_1$ is introduced at time zero and it finds a few specific *Ab*'s within the animal. After a lag phase, the *Ab* against antigen $Ag_1$ appears and its concentration rises up to a certain level and then starts to decline (*primary response*). Consider at this point the exposition to an antigen $Ag_2$ not correlated with antigen $Ag_1$. Then, no specific Ab is present and the Ab response will be similar to that obtained in the case of $Ag_1$ [13]. On the other hand, one important characteristic of the immune memory is that it is *associative*: B cells adapted to a certain type of antigen present a faster and more efficient secondary response not only to, but also to any structurally related antigen. This phenomenon is called *immunological cross reaction* or *cross-reactive response* [14]–[18]. This associative memory is contained in the process of vaccination and is called *generalization* capability or simply generalization in other artificial (computational) intelligence fields, like neural networks [19].

Some characteristics of the associative memories are particularly interesting in the context of AIS:
1) the stored pattern is recovered through the presentation of an incomplete or corrupted version of the pattern;
2) they are usually robust, not only to noise in the data, but also to failure in the components of the memory.

By comparison with the primary response, the secondary response is characterized by a shorter lag phase, a higher rate, and longer synthesis of Ab's with higher antigenic affinities (see the affinity maturation section). Moreover, a dose of Ag substantially lower than that required to initiate a primary response may cause a secondary response.

Under an engineering perspective, the cells with higher affinity must somehow be preserved as high-quality candidate solutions and shall only be replaced by improved candidates, based on statistical evidences. This is the reason why the AIS has a specific memory set as part of the whole repertoire.

As a summary, immune learning and memory are acquired through [19]:
1) repeated exposure to an Ag;
2) affinity maturation of the receptor molecules;
3) low-grade chronic infection;
4) cross-reactivity.

### B. Affinity Maturation

In an immune response, the repertoire of Ag-activated B cells is diversified basically by two mechanisms:
1) hypermutation and 2) receptor editing [20]–[23].

Ab's present in a memory response have, on average, a higher affinity than those of the early primary response. This phenomenon is referred to as the *maturation of the immune response*. This maturation requires the *Ag*-binding sites of the *Ab* molecules to be structurally different from those present in the primary response.

Random changes are introduced into the genes responsible for the *Ag*–*Ab* interactions and occasionally one such change will lead to an increase in the affinity of the *Ab*. These higher affinity variants are then selected to enter the pool of memory cells. Not only the repertoire is diversified through a *hypermutation mechanism*, but also mechanisms must exist such that rare B cells with high affinity mutant receptors can be selected to dominate the response. Those cells with low affinity or self-reactive receptors must be efficiently eliminated [20]–[22].

Somatic mutations guide to local optima, while receptor editing introduces diversity, leading to possibly better candidate receptors. In conclusion, point mutations are good for exploring local regions, while editing may rescue immune responses stuck on unsatisfactory local optima.

### III. AUTOMATED PROGRAM VERIFICATION AND PROGRAM INVARIANTS

The field of automated program verification started with seminal work by Floyd [24] and Hoare [25]. They introduced the concept of a *loop invariant*: a mathematical formula that remains true throughout the execution of a loop. The loop invariant completely captures the semantics of the loop, and along with the program preconditions and postconditions, can be used to show correctness of the program [25].

Recent work [8] has shown how the loop invariant for a particular program can be generated by *a priori* agreement on the *shape of the invariant*: the approximate polynomial representation of the invariant. However, the shape of the loop invariant can be hard to deduce for many programs.

The following shows an example program:

```
{A ≥ 0, B ≥ 0}
x := A;
y := B;
z := 0;

while x > 0 do
  if odd(x) then z := z + y;
```

```
    y := 2 * y;
    x := x/2;
  end while
```

Assuming the shape of the program invariant as $I_{shape}$: $Ax + By + Cz + Dxy + Eyz + Fxz + Gxyz + H = 0$, (where $A, B, C, D, E, F, G$ and $H$ are constants or program variables), using quantifier elimination [8], I get the final loop invariant as $I_{final}$: $z + xy - AB = 0$. Coupled with a precondition **P**: $\{A \geq 0 \wedge B \geq 0 \wedge x = A \wedge y = B \wedge z = 0\}$ and a postcondition **Q**: $\{z = A*B\}$, it can be shown that this invariant is consistent with **Q** i.e. the program correctly multiplies 2 numbers $A$ and $B$ and stores the result in $z$.

Finding the precise shape of the loop invariant is generally a non-trivial process and the AIS algorithm proposed aims to use "cues" from the program to make informed predictions about the invariant shape and ultimately help in automated program verification.

## IV. PROPOSAL

An AIS algorithm will be used to generate shapes of program invariant. Initially the AIS will be trained on programs, for which the shape of invariant is known. Then a program will be presented to the AIS and it will try to predict the form of the invariant.

An AIS approach presents many advantages over a traditional Machine Learning (ML) approach. In an AIS, recognition can be *sloppy* [26] i.e. if it has previously recognized program *P* (with an invariant *I*), then a new program *P'* "similar" to *P*, can also be recognized, and an invariant *I'* can be generated, that is similar in form to *I*. This is akin to our immune system recognizing a previously encountered pathogen (program), and generating antibodies (invariant) similar to the previously produced antibodies.

The natural immune system produces antibodies by a process of mutation, and the same process is emulated in AIS algorithms. A candidate solution (invariant) will be generated, and then the solution will be improved by *in-silico* mutation.

Previously encountered programs and their corresponding invariants will be stored as memory B cells. When a program similar to a stored one is presented, the time taken to generate the invariant will be shorter than the time taken to generate the original invariant (*secondary response*).

It is shown that such an evolutionary computing approach can help make more informed predictions about shapes of program invariants.

## V. COMPONENTS OF THE AIS

Having resolved the question of how automated program verification can benefit from an AIS approach, the specific components of the AIS have to be determined. What is the program analogue of an antigen and an antibody?

A *program fragment* is defined to be either an assignment statement, a statement containing an iteration construct (for, while, repeat, etc), or a statement having a conditional check (if <condition> then) e.g. *x := x + 2*, and *while (x > 0) do*, and *if (x > 3) then*, are all program fragments.

The analogue of an antigen is a program fragment and the corresponding analogue of an antibody is an invariant for the program fragment it recognizes. Hence, the AIS will be presented with an antigen (program fragment), and the immune system cells will either produce the antibody (invariant) immediately if it has encountered this antigen before, or will undergo somatic hypermutations to generate the correct antibody (invariant).

The individual invariants for each program fragment will then be recombined to generate the invariant for the whole program.

## VI. A SHAPE SPACE AND ANTIGENIC DISTANCE FOR PROGRAMS

We need a measure of distance between disparate program fragments, so that the AIS can recognize them and generate an antibody in response. For a natural immune system, the antibody combining region relevant to antigen binding can be specified by a number of "shape" parameters [27]. Some of these parameters would be geometric quantities specifying the size and shape of the combining site, and others would specify physical characteristics of the amino acids comprising the combining site such as charge, dipole moment, and the ability to form hydrogen bonds.

If there are N shape parameters, they can be combined into a vector, and antibody combining sites and antigenic determinants can be described as points **Ab** and **Ag**, in an N-dimensional Euclidean vector-space called *shape space* [27].

*Antigenic distance* between 2 antigens is the distance in shape space [14] between them e.g. $\|Ag_1 - Ag_2\|$ is the distance between antigens $Ag_1$ and $Ag_2$ in shape space S. The *antibody distance* is the distance $\|Ab_1 - Ab_2\|$ in shape space between 2 antibodies $Ab_1$ and $Ab_2$.

I define the *program fragment shape space* as the N-dimensional Euclidean vector space of program fragment characteristics like identifier name, exponent on the identifier, operator, etc. I define the corresponding *program fragment antigenic distance* as the distance $\|P_1 - P_2\|$ between 2 program fragments $P_1$ and $P_2$ in program fragment shape space. The *program fragment antibody (invariant) distance* is the distance $\|I_1 - I_2\|$ between 2 program fragments $I_1$ and $I_2$ in program fragment shape space.

Let us consider 2 program fragments $P_1$: *x := x + 2* and $P_2$: *t := t + 2*. The corresponding antibody (invariant) for $P_1$ is $I_1$: $x = x + 2n$, where *n* is a program variable or constant (since upon *n - 1* iterations, *x* gets the value *x + 2n*). Let $P_1$ and $I_1$ constitute the training set. Then the AIS should be able to produce an antibody (invariant) for the program fragment $P_2$ even though it has never encountered this antigen (program) before. The correct invariant is $I_2$: $t = t + 2n$ (where *n* is a program variable or constant) and this is indeed what the AIS generates by somatic hypermutation. The program $P_1$ differs from $P_2$ by 1 mutation (replacing *x* by *t* on both sides of the assignment) i.e. the *program fragment antigenic distance* ‖**P₁** - **P₂**‖ is 1. The invariants **I₁** and **I₂** also differ by 1 mutation (replacing *x* by *t*) i.e. the *program fragment antibody (invariant) distance* ‖**I₁** - **I₂**‖ is 1. Hence, when an AIS trained on (**P₁**, **I₁**) is presented with **P₂**, it produces **I₂** using one mutation from **I₁** (Fig. 1).

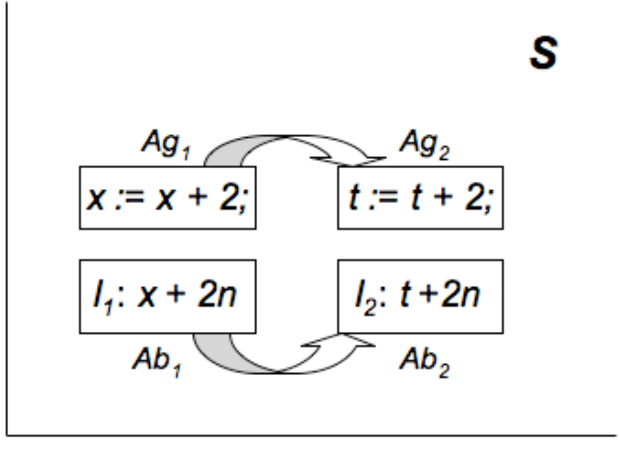

Fig. 1. AIS mutation from the assignment statement **Ag₁** (`x := x + 2;`) and invariant **Ab₁** (`x + 2n`) to **Ag₂** (`t := t + 2;`) and invariant **Ab₂** (`t + 2n`) in shape space S.

## VII. ALGORITHM

The AIS was implemented in MATLAB® and trained on the antigen (program fragment) **P₁**: $x := x + 2$ and given the antibody (invariant) **I₁**: $x = x + 2n$ as a solution (*training phase*). The AIS stores the solution **I₁** as a memory detector.

When an entire program (as opposed to a program fragment) is presented to the AIS, it breaks the program up into program fragments (all the assignment statements in the program), and then "presents" each of these antigens (fragments) to itself.

If an antigen (program fragment) **P₂** "similar" to **P₁** is detected, it will generate **I₁** as a candidate solution. If **I₁** itself does not act as an invariant, the AIS will keep on carrying out mutations on **I₁** until it evolves the final antibody (invariant) **I₂** that will act as the invariant for the program presented (*somatic hypermutation phase*). This is akin to how the natural immune system mutates B cell receptors (BCR) and ultimately produces a BCR that can recognize the antigen. In the last step, the AIS incorporates **Iᵢ** into its memory pool (*learning*).

The AIS then presents the next program fragment **P₃**, generates the invariant **I₃** and stores it in the memory population, and so on until all program fragments have been presented. Finally, the AIS combines all invariants linearly, producing a polynomial (shape of invariant) that captures the semantics of the entire program.

## VIII. RESULTS

The AIS (trained on **P₁**, **I₁**) was presented with suites of entire programs and it successfully generated the shape of their invariant. The first program is shown below:

```
(x,y,u,v) := (a,b,b,0);
x := a; y := b;
u := b; v := 0;

while (x ≠ y) do
  while (x > y) do x := x - y; v := v + u;
  end while;
  while (x < y) do y := y - x; u := u + v;
  end while;
end while
```

This program takes 2 positive integers *a* and *b*, and calculates their g.c.d and l.c.m. The AIS presents itself with each assignment statement sequentially. The first 4 assignment statements (lines 1-2) have no invariant, since they are not contained inside any loop. Hence, the AIS does not generate any invariant for them. The progress of the algorithm on the next 2 assignment statements (`x := x - y; v := v + u;`) is shown below:

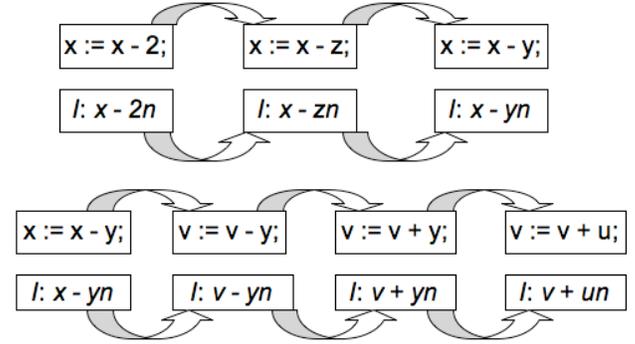

Fig. 2. AIS mutations for the assignment statements `x := x - y; v := v + u;`

The AIS starts from the training set (**P₁**: $x := x + 2$ & **I₁**: $x = x + 2n$) and then mutates the operators and operands to create the invariant **I₃**: $x = x - yn$ for the program fragment **P₃**: $x := x - y$. The AIS stores **I₃** in the memory population and for the next assignment statement (`v := v + u;`), it starts mutating from (**P₃**,**I₃**) until it creates the invariant **I₄**: $v = v + un$ for the program fragment **P₄**: $v := v + u$.

For the next set of assignment statements (`y := y - x; u := u + v;`), the AIS then generates the invariants **I₅**: $y = y - xn$ and **I₆**: $u = u + vn$ (not shown). The 4 invariants **I₁**, **I₂**, **I₃** & **I₄** are then combined linearly (with *n* being substituted for all program variables, namely *x, y, u, v*) to yield the invariant shape **I**$_{shape}$: $Ax + By + Cy + Du + Exy + Fy^2 + Guy + Hvy + Jxu + Ku^2 + Lvu + Mx^2 + Nvx + Pv^2 + Q = 0$, where *A, B, C, D, E, F, G, H, J, K, L, M, N, P* and *Q* are constants or program variables. This is the correct invariant shape, since using quantifier elimination [8], I get the final invariant **I**$_{final}$: $xu + yv - ab = 0$ (with $A = B = C = D = E = F = G = K = L = M = N = P = 0$, $Q = -ab$, $H = J = 1$).

The AIS was also tested on another standard program [8] shown below:

```
{A ≥ 0, B ≥ 0}
x := A;
y := B;
z := 1;

while y > 0 do
if odd(y) then y := y - 1;  z := x * z;
else x := x * x;   y := y/2;
end while
```

This program calculates $A^B$ and stores it in $z$. The AIS correctly calculates the invariant for the program fragment $P_5$: $z := x * z$ as $I_5$: $z = x^n * z$. For the program fragment $P_6$: x := x * x, it generates the invariant $I_6$: $x = exp(x, exp(2,n))$, where $exp()$ is the exponentiation function. Combining all the program fragment invariants, gives us the following invariant shape:

$I_{shape}$: $Azx^x + Bzx^y + Czx^z + D.exp(x,exp(2,x)) + E.exp(x,exp(2,y)) + F.exp(x,exp(2,z)) + G = 0$.

This is the exact shape of the invariants, since quantifier elimination yields the final invariant

$I_{final}$: $zx^y = A^B$ (with $A = C = D = E = F = 0$, $G = -A^B$).

We can now readily verify the working of the program. When the loop terminates, the invariant is true and $y = 0$, which yields the correct postcondition: $z = A^B$.

## IX. CONCLUSION AND FUTURE WORK

I have presented an immune system inspired approach for automated program verification. The AIS algorithm breaks up a program into fragments and presents them to itself. It then generates an invariant in response to each program fragment and ultimately combines them to create the general shape of the invariant. This AIS algorithm was tested on non-trivial benchmark programs [8] and was found to correctly generate the general form of the program invariant.

Future work will focus on theoretical research into whether there are classes of programs for which a linear combination of individual program fragment invariants might not generate the invariant for the entire program. Research is also ongoing into how mutations on exponentiation would affect the invariant e.g. *x := x + 2* getting mutated to *x := x$^2$ + 2*. Lastly, the AIS algorithm does not consider program fragments having iteration constructs like *while*, *repeat*, etc. and future research will investigate how incorporation of such program fragments can enhance the predictive power of the algorithm.

ACKNOWLEDGMENTS

The author wishes to thank Prof. Deepak Kapur and ThanhVu Nguyen for helpful comments.